# AUTONOMOUS CRATER DETECTION ON ASTEROIDS USING A FULLY-CONVOLUTIONAL NEURAL NETWORK

F. Latorre[1*], D. Spiller[2], F. Curti[1]

[1]School of Aerospace Engineering, Sapienza University of Rome, Via Salaria 851, 00138 Rome, Italy

[2]Italian Space Agency, Via del Politecnico snc, 00133 Rome, Italy

* latorre.1597476@studenti.uniroma1.it

**ABSTRACT**

*This paper shows the application of autonomous Crater Detection using the U-Net, a Fully-Convolutional Neural Network, on Ceres. The U-Net is trained on optical images of the Moon Global Morphology Mosaic based on data collected by the LRO and manual crater catalogues. The Moon-trained network will be tested on Dawn optical images of Ceres: this task is accomplished by means of a Transfer Learning (TL) approach. The trained model has been fine-tuned using 100, 500 and 1000 additional images of Ceres. The test performance was measured on 350 never before seen images, reaching a testing accuracy of 96.24%, 96.95% and 97.19%, respectively. This means that despite the intrinsic differences between the Moon and Ceres, TL works with encouraging results.*
*The output of the U-Net contains predicted craters: it will be post-processed applying global thresholding for image binarization and a template matching algorithm to extract craters positions and radii in the pixel space. Post-processed craters will be counted and compared to the ground truth data in order to compute image segmentation metrics: precision, recall and F1 score.*
*These indices will be computed, and their effect will be discussed for tasks such as automated crater cataloguing and optical navigation.*

## 1 INTRODUCTION

Crater Detection Algorithms (CDAs) are traditionally used to automate the process of crater counting for cataloguing means. These algorithms typically rely on classical machine learning [1][2], neural networks [3]. In particular, Convolutional Neural Networks (CNNs) [4][5][6][7], learn how to extract image features on their own. CNNs can be used for classification and segmentation tasks: segmentation networks [8] are Fully-Convolutional Neural Networks (FCN), because they take an arbitrary image (with minimal pre-processing) as input and give a segmented image as output, resulting in a very accurate performance. One of the most famous FCNs, the U-Net [9] was initially thought for biomedical image segmentation of biological cells, and later adapted to crater detection and counting on the Moon [10] and Mars [11][12].

Crater Detection for Terrain Relative Navigation and hazard avoidance has been proposed using intensity (i.e. in the visible spectrum) images for detection and spacecraft state estimation with an Extended Kalman Filter [13]. In general, an on-board implementation could include optical (passive) or radar (active) measurements. Optical measurements are smaller, but more prone to sunlight variation because the crater's shade position can change during times of the day; radar measurements are more robust to sunlight variation, because they rely on elevation measurements instead: however, they are more complex.

The U-Net is trained on a total of 30,000 images of the Moon, validated on 5,000 images and tested on 5,000 images, all taken from the LRO Global Morphology Mosaic [14] of the Moon. The trained model will then be transferred and fine-tuned on data coming from a Ceres Mosaic for testing purposes. The choice of this particular asteroid depends on the fact that it has an already existing reference catalogue which can be used both for fine-tuning and crater counting.

Three different cases, in which 100, 500 and 1,000 images are used as additional training data, are discussed: different results are obtained in terms of training accuracy and crater counting performance. The raw results, obtained on 350 test images, will be post-processed using global thresholding and passed through a template matching algorithm to extract the craters' positions and radii in pixel space. Visual results will then be inspected and discussed. The resulting found craters will be counted and compared with a reference catalogue using the same template matching algorithm mentioned before, leading to computation of the segmentation metrics of precision, recall and F1 score for all the test cases. These results will be compared each other and discussed for a possible application for navigation purposes.

The main goals of this work are summed up as follows:

- Training a U-Net on a Moon crater dataset and apply transfer learning (TL) on Ceres
- Validating transfer learning despite the differences between the original and new dataset.
- Applying segmentation for Crater Detection on asteroids for the first time.
- Discussing the possibility of implementation of Crater Detection for Terrain Relative Navigation.

## 2 IMAGE SEGMENTATION

### 2.1 U-Net

Figure 1 shows the architecture of the U-Net.

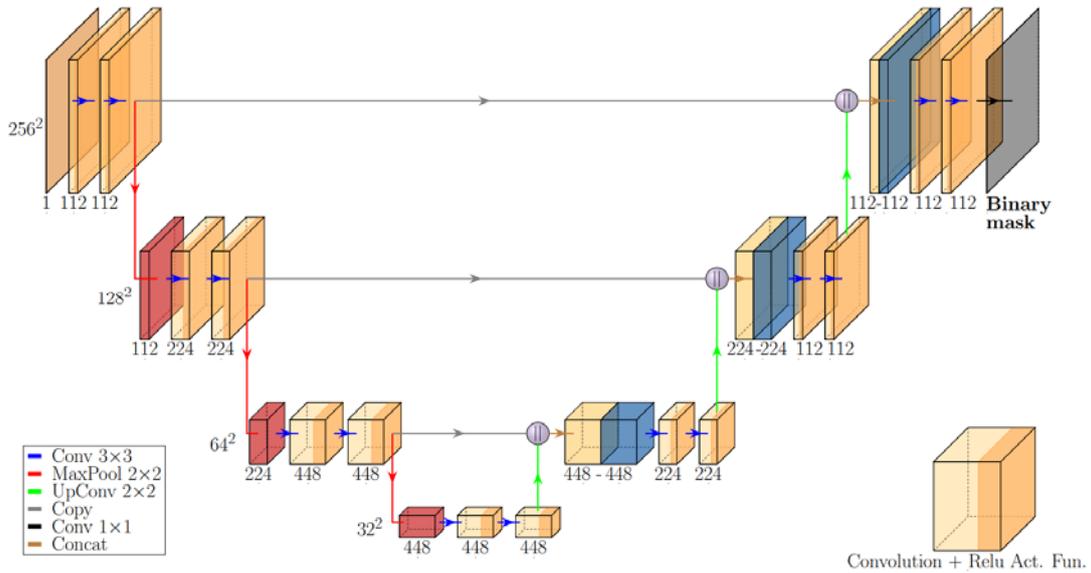

Figure 1: U-Net model with features dimensions and number of features.

### 2.2 Segmentation metrics

The group of accuracy metrics that has been used the most for crater detection with CNNs consists of precision, recall and F1 score. Precision tells how many matches are detected with respect to all the craters found by the network:

$$P = \frac{T_P}{T_P + F_P} \quad (1)$$

Recall tells how many matches are detected with respect to the annotated craters, so it gives an idea of how many already existing craters are found by the network:

$$R = \frac{T_P}{T_P + F_N} \quad (2)$$

True positives ($T_P$) are matches between found and ground truth craters.
False positives ($F_P$) are craters found by the CNN which are not present in the ground truth.
False negatives ($F_N$) are craters missed by the network but contained in the ground truth.
When high precision is reached, recall is slightly penalized and vice versa. For this reason, a better score could be the harmonic mean between precision and recall, which is also called F1 score:

$$F_1 = \frac{2PR}{P + R} \quad (3)$$

## 3   TRANSFER LEARNING ON CERES

Transfer learning (TL) is a term that refers to the capability to infer information about a new domain after training on a different domain.
TL was previously applied using a U-Net fed with Mars [11] and Mercury [10][16] DEM images after training the network with lunar data; in this paper, it will be taken into consideration for asteroids, namely Ceres.

### 3.1   Mathematical notation

A domain $\mathcal{D}$ is defined as:

$$\mathcal{D} = \{\mathcal{X}, P(X)\} \quad (4)$$

Where $\mathcal{X}$ is the feature space of a specific domain, and $P(X)$ is a marginal probability distribution, being $X = x_1, \ldots, x_n \in \mathcal{X}$.
In transfer learning two different domains can be defined: a source domain $\mathcal{D}_S$ and the target domain $\mathcal{D}_T$. Two domains can be different either if $\mathcal{X}_T \neq \mathcal{X}_S$ or $P(X_T) \neq P(X_S)$.
A task $\mathcal{T}$ is defined as

$$\mathcal{T} = \{\mathcal{Y}, P(Y \mid X)\} \quad (5)$$

Where $\mathcal{Y}$ is the label space, $P(Y \mid X)$ is the conditional probability distribution, which is the probability that $y \in \mathcal{Y}$ is inferred from $x \in \mathcal{X}$. Following the same notation used for the domains, two tasks $\mathcal{T}_S$ and $\mathcal{T}_T$ can be defined for the source and target domain, respectively. Two tasks can be different either if $\mathcal{Y}_T \neq \mathcal{Y}_S$ or $P(Y_T \mid X_T) \neq P(Y_S \mid X_S)$. All the combinations lead to four different scenarios:

- $\mathcal{X}_T \neq \mathcal{X}_S$
- $\mathcal{Y}_T \neq \mathcal{Y}_S$
- $P(X_T) \neq P(X_S)$
- $P(Y_T \mid X_T) \neq P(Y_S \mid X_S)$

The first two scenarios are commonly referred to as heterogeneous transfer learning, which occurs if features and/or labels are different in the two domains (but probability distributions are not).
On the other hand, the last two scenarios are grouped together as homogeneous transfer learning, in which features and labels are the same in the two domains, but their marginal and/or conditional probabilities are different.

This work will take into consideration a homogeneous TL: in fact, feature spaces are images labelled with masks containing either craters or non-craters.

### 3.2 Moon training

Before being fed with Ceres data, the U-Net has to be pre-trained in order to obtain the model upon which transfer learning can be applied. The network is trained upon the Moon Global Mosaic and tested on a Ceres Global Mosaic. This means that no ambiguity or critical issues for transfer learning generalization are expected, because the data type has the same nature. The intrinsic differences between the contexts of the Moon and the asteroids, however, may lead to domain-specific crater shapes, sizes and frequency, so their marginal and conditional probabilities could be different as well.
The dataset used in this work is made of 256x256 images taken from the Lunar Recoinnassance Orbiter (LRO) [14], spanning the whole surface. Data has a resolution of 303 pixel per degree (or 100 meters per pixel).
The corresponding binary ground truth masks for each image are created using a combination of Head [16] and Povilaitis [17] catalogues as a source.

### 3.3 Ceres dataset and catalogue

The Ceres Global Mosaic was constructed based on data collected during the Dawn spacecraft mission [18], and it is available online. The representation for this asteroid is in a Plate Carree projection based on Framing Camera (FC) images and covers the whole asteroid surface. The resolution of the mosaic is 140 meters per pixel (58 pixels per degree).
The reference catalogue is Zeilnhofer [19], built after analysis of data taken during the LAMO (Low Altitude Mapping Orbit) phase of the spacecraft, at an altitude of 850 km.

### 3.4 Pre-processing methods: image cropping and orthographic projection

The Global Mosaic has to be cropped into 256x256 images, which are later orthographically projected to prevent crater warping at high latitudes.
Given the geographic coordinate limits of the source and of the wanted output image $c_i = [c_{ix,min}, c_{ix,max}, c_{iy,min}, c_{iy,max}]^T$ and $c_o = [c_{ox,min}, c_{ox,max}, c_{oy,min}, c_{oy,max}]^T$, respectively, along with its width $w$ and height $h$, the transformation into pixel coordinates is:

$$x = \begin{bmatrix} x_1 \\ x_2 \end{bmatrix} = \frac{w}{c_{ix,max} - c_{ix,min}} \left( \begin{bmatrix} c_{ox,min} \\ c_{ox,max} \end{bmatrix} - c_{ix,min} \begin{bmatrix} 1 \\ 1 \end{bmatrix} \right) \tag{6}$$

$$y = \begin{bmatrix} y_1 \\ y_2 \end{bmatrix} = \frac{w}{c_{iy,max} - c_{iy,min}} \left( \begin{bmatrix} -c_{oy,min} \\ -c_{oy,max} \end{bmatrix} + c_{iy,max} \begin{bmatrix} 1 \\ 1 \end{bmatrix} \right) \tag{7}$$

$x_1, y_2, x_2, y_1$ are the pixel coordinate limits for image cropping. To prevent crater warping at high latitudes in the Plate Carree representation, the cropped images are orthographically projected [20].

### 3.5 Post-processing methods

The raw U-Net predictions are binarized by means of a global thresholding.
The term thresholding refers to image binarization with respect to some fixed threshold value T. Pixels above this value are put to 1, while pixels below this value are put to 0. So, being $f(x, y)$ and $g(x, y)$ the original and the binarized image, respectively, the thresholding operation can be defined as follows:

$$g(x, y) = \begin{cases} 1 & if\ f(x, y) > T \\ 0 & if\ f(x, y) < T \end{cases} \tag{8}$$

### 3.5.1 Template matching

After image binarization, a template matching algorithm is applied. The algorithm finds the most probable coordinates (x,y,r), in pixel space, of the craters, where (x,y) is the centroid of the ring, and r is its radius. Templates are built choosing a minimum and maximum radius value.
A circle is detected if the correlation between the generated template and the target is above a certain probability value.
The template matching algorithm also serves as a tool to discard duplicate craters. Duplicate craters can occur because of random image sampling: this could mean that the same crater can appear inside two different images, in another location. Two craters i and j that satisfy the following criteria are considered as duplicates:

$$\frac{(x_i - x_j)^2 + (y_i - y_j)^2}{\min(r_i, r_j)^2} < D(x, y) \qquad (9)$$

$$\frac{r_i - r_j}{\min(r_i, r_j)} < D_r \qquad (10)$$

where $D(x, y)$ and $D_r$ are thresholding parameters for the pixel coordinates and the radius.

### 3.5.2 Crater matching

To compute matches between detected craters and ground truth data (true positives), template matching is used once again to extract the craters' positions and radii, which are then compared to the corresponding ground truth. Similarly to Equations (9) and (10), a match between ground truth and prediction is defined if the following two criteria are fulfilled:

$$\frac{(\phi_{ref} - \phi)^2 + (\lambda_{ref} - \lambda)^2}{\min(R_{ref}, R)^2} < D_{x,y} \qquad (11)$$

$$\frac{R_{ref} - R}{\min(R_{ref}, R)} < D_r \qquad (12)$$

The variables $(\phi_{ref}, \lambda_{ref}, R_{ref})$ are the pixel coordinates of the ground truth crater, $(\phi, \lambda, R)$ are the pixel coordinates of the predicted crater. These variables are referred to the crater's center latitude, longitude and radius. $D_{x,y}$ and $D_r$ are the same user-defined threshold parameters in template matching.

## 4    RESULTS

### 4.1    Moon training

For the network training on the Moon, 30,000 images were chosen for training, 5,000 for validation and 5,000 for testing, all with size 256x256. An example of a Moon image with the associated crater mask is shown in Figure 2.

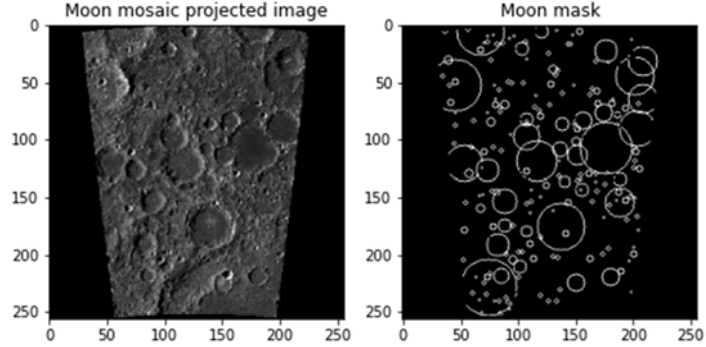

Figure 2: Sample of Moon mosaic image and its corresponding mask used for the U-Net training.

The loss function to be minimized during the training is a pixel-wise binary cross entropy, which is used for segmentation problems,

$$l_i = x_i - x_i z_i + \log(1 + e^{-x_i}) \tag{15}$$

Where $x_i$ is the predicted value and $z_i$ is the ground truth pixel value. The training was done using TensorFlow Keras [21], on an NVIDIA GeForce RTX 2060 GPU.
The network was compiled with an ADAM optimizer [22], with added regularization and dropout to prevent overfitting. The training parameters are summed up in Table 1, and the results of the training are shown in Table 2. Weights obtained after the Moon training are saved and used for later transfer learning.

| Parameter | Value(s) |
|---|---|
| Learning rate | $10^{-4}$ |
| Regularization | $10^{-5}$ |
| Dropout | 0.15 |
| Epochs | 30 |
| Batch size | 3 |

Table 1: U-Net parameters settings.

|  | Training | Validation | Testing |
|---|---|---|---|
| Loss | 0.0988 | 0.1001 | 0.0988 |
| Accuracy | 96.19% | 96.07% | 96.09% |

Table 2: U-Net Moon training results.

### 4.2 Ceres fine-tuning

A varying number of images were cropped from the Global Mosaic and projected using Cartopy [23], with corresponding masks (Figure 3). The number of cropped images depends on how many of them will be used during fine-tuning of the model. In this work, 100, 500 and 1,000 images were chosen for three test cases and used to fine-tune the Moon-trained model for 10 epochs. The results are shown in Table 3.

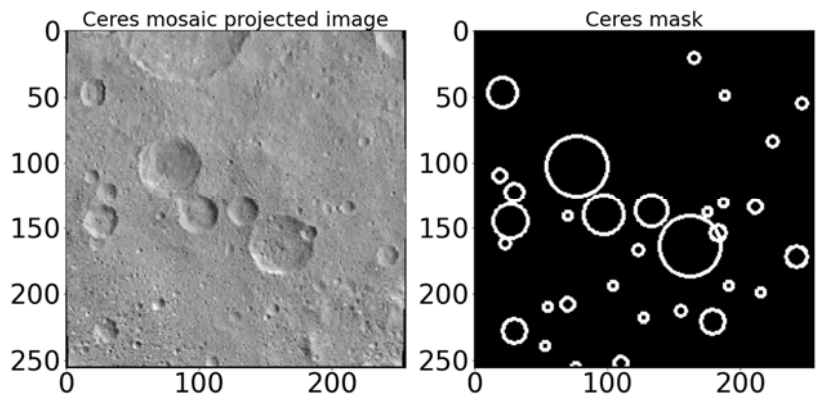

Figure 3: Sample of Ceres mosaic image used for Transfer Learning.

| *Training images* | **100** | **500** | **1,000** |
|---|---|---|---|
| Training accuracy | 97.54% | 98.42% | 98.81% |
| Testing accuracy | 96.24% | 96.95% | 97.19% |

Table 3: Fine-tuning and Transfer Learning results.

### 4.3 U-Net predictions

The raw outputs obtained from the U-Net show that crater detection works, as comparison between the ground truth and prediction shows. Figures 4 shows the prediction results for 100, 500 and 1,000 additional Ceres images, respectively. It can be seen that feeding that network with an increasing number of additional images for fine-tuning, the crater rims appear to be more clear and almost overlap with their corresponding ground truth masks. However, especially with just a few training images, some crater rims in the raw predictions could appear to be faint or oddly shaped, so additional post-processing has to be thought before feeding the predictions to the final crater matching algorithm. Post-processing is done through thresholding and template matching and its results will be shown next.

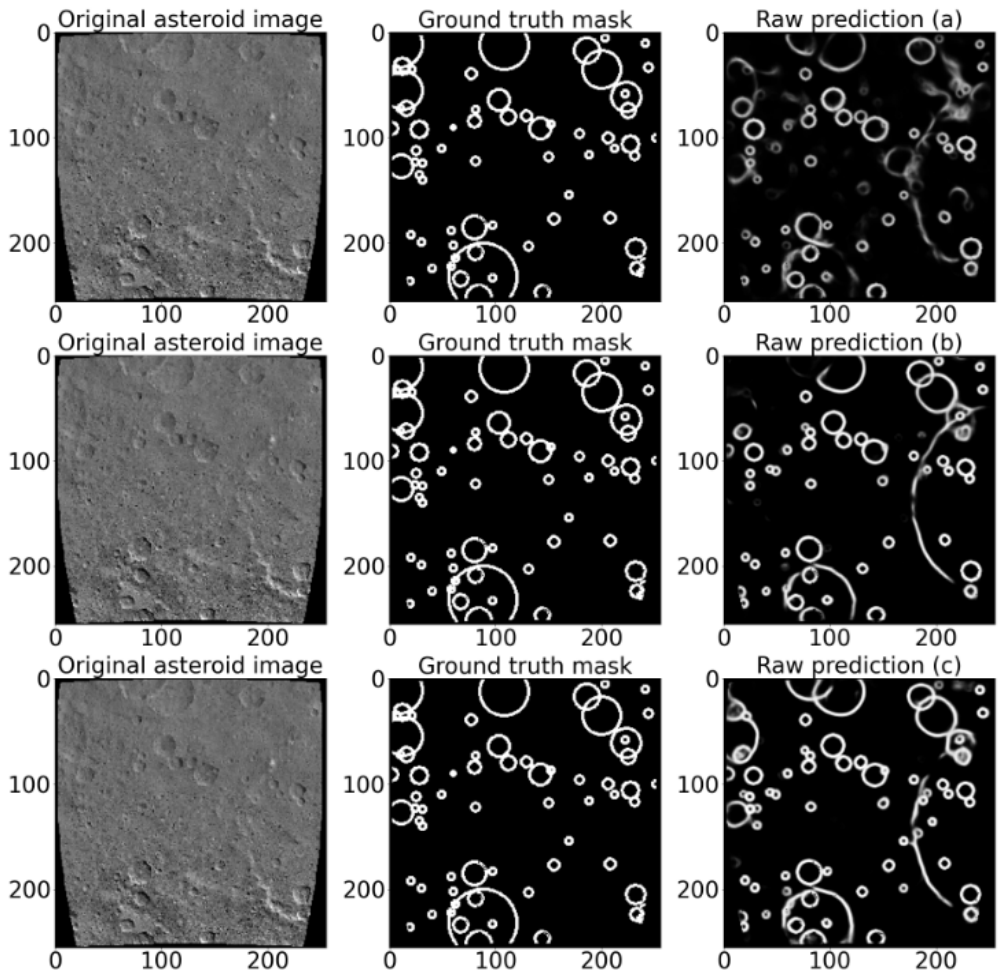

Figure 4: Comparison between ground truth and raw predictions. (a) Using 100 Ceres images. (b) Using 500 Ceres images. (c) Using 1,000 Ceres images.

### 4.4 Ceres post-processing

The thresholding was applied using Skimage [24] with a fixed threshold value T=0.1.
After the images were thresholded, OpenCV [25] template matching is applied: its parameters are summed up in Table 4.
The algorithm results, as shown in Figure 5, are then discriminated comparing them to the ground truth, following the method described before.

| Parameter | Value |
|---|---|
| $r_{min}$ | 5 |
| $r_{max}$ | 40 |
| $P_m$ | 0.5 |
| $D_{x,y}$ | 1.8 |
| $D_r$ | 1 |
| $T$ | 0.1 |

Table 4: Template matching parameters.

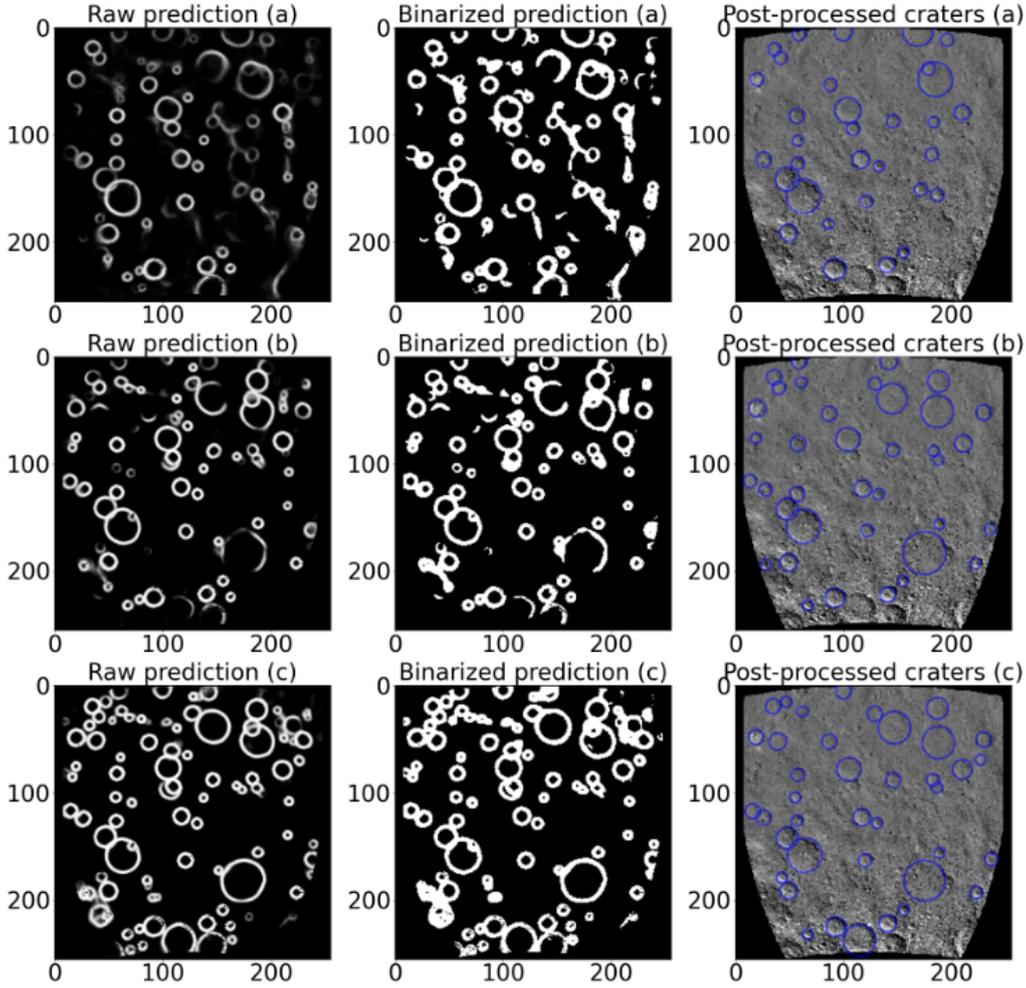

Figure 5: Ceres raw prediction, thresholded prediction and post-processed mask. (a) Using 100 Ceres images. (b) Using 500 Ceres images. (c) Using 1,000 Ceres images.

The total number of craters for each test case was detected by the U-Net over 350 test images sampled from the global mosaic. The segmentation performance was analysed considering precision, recall and F1 score. These metrics are summed up in Table 5 and later discussed.

| *Training images* | **100** | **500** | **1,000** |
|---|---|---|---|
| **Precision** | 70.51% | 84.43% | 83.45% |
| **Recall** | 48.27% | 53.32% | 58.97% |
| **F1** | 57.30% | 65.36% | 69.11% |

Table 5: Crater matching results.

## 5    DISCUSSION

The results presented in the last section show the application of a fine-tuning strategy for crater detection on Ceres, with variable outcomes.
This strategy, which takes the previously trained U-Net, and additionally trains all the layers in the network, reaches a good performance in terms of network accuracy, with maximum overfitting of 1.62%. Visual inspection shows that most of the ground truth craters are recovered by the network. In fact, this is confirmed by values obtained by the precision metric, which reaches a percentage of 84.43% after fine-tuning with 500 images. A high precision score means that a high percentage of true positives is found with respect to all the craters found by the network, including all the false positives.

This can be interpreted as minimization of the number of false positives, meaning that the U-Net can recover a significant amount of craters already stored in the catalogue. When precision reaches these values, recall is automatically penalized: in fact, it reaches at most approximately 60% after fine-tuning with 1,000 images. A low recall corresponds to a high number of false negatives, so the network misses some craters that are expected to be found, as they are stored in the catalogue. Precision and recall alone do not allow an accurate analysis of the segmentation performance quality, especially relative to other network performances for the same task. The F1 score is a trade-off between precision and recall, and this makes it a good performance measure for comparison with other works in literature. In this paper, a F1 of 57.30%, 65.36% and 69.11% has been reached for the 100, 500 and 1,000 images, respectively. The results obtained using 500 and 1,000 images especially are comparable with DeepMoon, which uses the same training and post-processing parameters presented here and reaches an optimal F1 of 67% [10].

This paper also proposes a possible on-board implementation of this application, following this strategy:

- Train the U-Net offline with lunar data.
- Implement the U-Net with its pre-trained weights and the asteroid catalogue on-board.
- Scan the asteroid surface to obtain additional data to be fed to the U-Net for further training and fine-tuning.

## 6      CONCLUSION

In this paper, it was demonstrated that transfer learning worked even if the starting planetary body, the Moon, has different geological features and dimensions with respect to Ceres, an asteroid. Also, despite the image resolution differences, the U-Net reached a good training and testing accuracy and it was capable of recovering the most evident craters, even before post-processing application.

Image segmentation metrics have been computed on Ceres, because a detailed catalogue of its craters is available. As expected, the performance outcome grows with the number of images given to the network for additional training. The main drawback behind the network accuracy is that overfitting occurs, and it also grows with the number of images: this problem can be avoided by performing data augmentation (i.e. random image transformations such as rotation and translation). This is left for future analysis. As for what concerns segmentation metrics, precision is the preferred one, meaning that the recovered craters are mostly catalogued ones, but at the same time a significant amount of them is missed and many false negatives are found. Future work can be done in order to make recall higher: this is important in a crater detection context, where the false negatives number has to be kept as low as possible, since it is a missed crater which could lead to mission failure. This task can be accomplished by trying different network architectures and hyperparameters.

Despite all these flaws, these results are still encouraging, and for this reason they are considered as a step for a possible on-board implementation.

A similar analysis can be done on other asteroids such as Vesta, for which a global mosaic is available. However, up to now, only visual results may be analysed, because a catalogue is not available for comparison between predictions and ground truth.